\documentclass[10pt,twocolumn,twoside] {IEEEtran}
\usepackage{mathrsfs}
\usepackage{tabularx,booktabs}
\usepackage[cmex10]{amsmath}

\usepackage{multirow}
\usepackage{graphicx}
\usepackage{subcaption}
\usepackage{lineno}
\usepackage{setspace}
\usepackage{url}
\usepackage{amssymb}
\usepackage{color}
\usepackage[autostyle=false, style=english]{csquotes}
\MakeOuterQuote{"}

\def\eg{{\em e.g.}}
\def\ie{{\em i.e.}}

\usepackage{mathrsfs}

\def\m{{\bf m}}

\def\0{{\bf 0}}
\def\1{{\bf 1}}

\def\eg{{\em e.g.}}
\def\ie{{\em i.e.}}

\usepackage[pagebackref=false,breaklinks=true,letterpaper=true,colorlinks,bookmarks=false]{hyperref}
\usepackage[linesnumbered,ruled,vlined]{algorithm2e}
\graphicspath{{Figures/}}

\begin{document}
\title{Automatic deep learning-based normalization of breast dynamic contrast-enhanced magnetic resonance images}

\author{Jun~Zhang,
	Ashirbani~Saha,
	Brian J.~Soher,
	Maciej A. Mazurowski$^\dagger$	
	\thanks{$^\dagger$Corresponding author. }
	\thanks{J.~Zhang, A.~Saha, B. J. Soher, M. A.~Mazurowski are with Department of Radiology, Duke University, Durham, NC, USA (Email: maciej.mazurowski@duke.edu)}	
}
\maketitle
\begin{abstract}
\textit{Objective:} To develop an automatic image normalization algorithm for intensity correction of images from breast dynamic contrast-enhanced magnetic resonance imaging (DCE-MRI) acquired by different MRI scanners with various imaging parameters, using only image information. \textit{Methods:} DCE-MR images of 460 subjects with breast cancer acquired by different scanners were used in this study. Each subject had one T1-weighted pre-contrast image and three T1-weighted post-contrast images available. Our normalization algorithm operated under the assumption that the same type of tissue in different patients should be represented by the same voxel value. We used four tissue/material types as the anchors for the normalization: 1) air, 2) fat tissue, 3) dense tissue, and 4) heart. The algorithm proceeded in the following two steps: First, a state-of-the-art deep learning-based algorithm was applied to perform tissue segmentation accurately and efficiently. Then, based on the segmentation results, a subject-specific piecewise linear mapping function was applied between the anchor points to normalize the same type of tissue in different patients into the same intensity ranges. We evaluated the algorithm with 300 subjects used for training and the rest used for testing. \textit{Results:} The application of our algorithm to images with different scanning parameters resulted in highly improved consistency in pixel values and extracted radiomics features. \textit{Conclusion:} The proposed image normalization strategy based on tissue segmentation can perform intensity correction fully automatically, without the knowledge of the scanner parameters. \textit{Significance:} We have thoroughly tested our algorithm and showed that it successfully normalizes the intensity of DCE-MR images. We made our software publicly available for others to apply in their analyses.

\end{abstract}

\begin{IEEEkeywords}
Image intensity normalization, tissue segmentation, deep learning
\end{IEEEkeywords}

\section{Introduction}
Recent studies have demonstrated that dynamic contrast-enhanced magnetic resonance imaging (DCE-MRI) provides a useful tool for computer-aided breast cancer diagnosis~\cite{mahrooghy2015pharmacokinetic}, prognosis~\cite{mazurowski2015recurrence}, and correlation with genomics~\cite{mazurowski2015radiogenomics} (\ie, radiogenomics), thanks to both anatomical structures and physiological tissue characteristics contained in DCE-MRI~\cite{karahaliou2014assessing}.

A common challenge in DCE-MRI studies as well as MRI studies in general is that the images might have different intensity ranges, making the quantitative analysis difficult. This is primarily caused by different acquisition parameters of the MRI scanners~\cite{o2011dynamic}. A simple solution to deal with this challenge is image intensity normalization, \eg, by linearly stretching image intensities from the minimum value to the maximum value. However, due to the subject-specific characteristics of different images, such simple strategy will generate inconsistent intensity ranges for different images, and thus these normalized images cannot be directly compared. A recent study showed that image acquisition parameters such as the magnet strength and slice thickness have a significant impact on radiomics features extracted from DCE-MR images rendering the diagnostic/prognostic tasks extremely difficult~\cite{saha2017effects}.

Several advanced methods have been proposed for MR image intensity correction based on imaging parameters~\cite{nyul1999standardizing,collewet2004influence,studholme2004accurate,vovk2007review,weisenfeld2004normalization}. Most of the methods construct a physical model using imaging parameters to normalize different images into a common intensity space. However, existing methods for image intensity correction are \textit{not only} error-prone \textit{but also} difficult to generate robust results, because there are many parameters involved in constructing accurate physical models. Thus, a parameter-free method for image intensity correction is highly desired for radiogenomics analysis, to make all images comparable after normalization.

\begin{figure}[tp]
	\centering
	\includegraphics[width=0.5\textwidth]{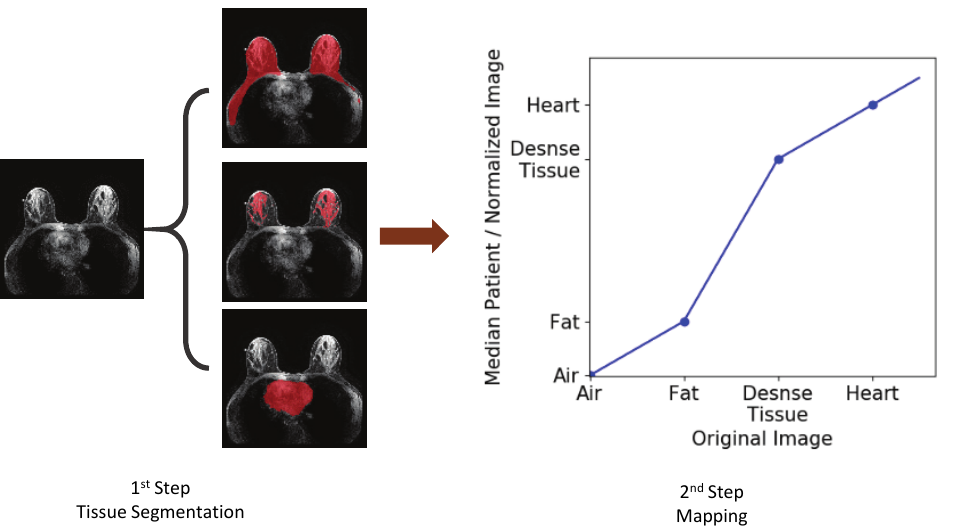}
	\caption{Pipeline of our proposed method. In the first step, we perform tissue segmentation (of air, fat tissue and dense tissue) for the input DCE-MR image, using a deep-learning algorithm. In the second step, based on the segmentation results, we perform a subject-specific piecewise linear transformation to normalize the same types of tissue in different patients into the same intensity range.}
	\label{outline}
\end{figure}

\begin{figure*}[!t]
	\setlength{\intextsep}{2pt plus 1pt minus 0pt}
	\setlength{\abovecaptionskip}{-1pt}
	\setlength{\belowcaptionskip}{-1pt}	
	\centering
	\includegraphics[width=\textwidth]{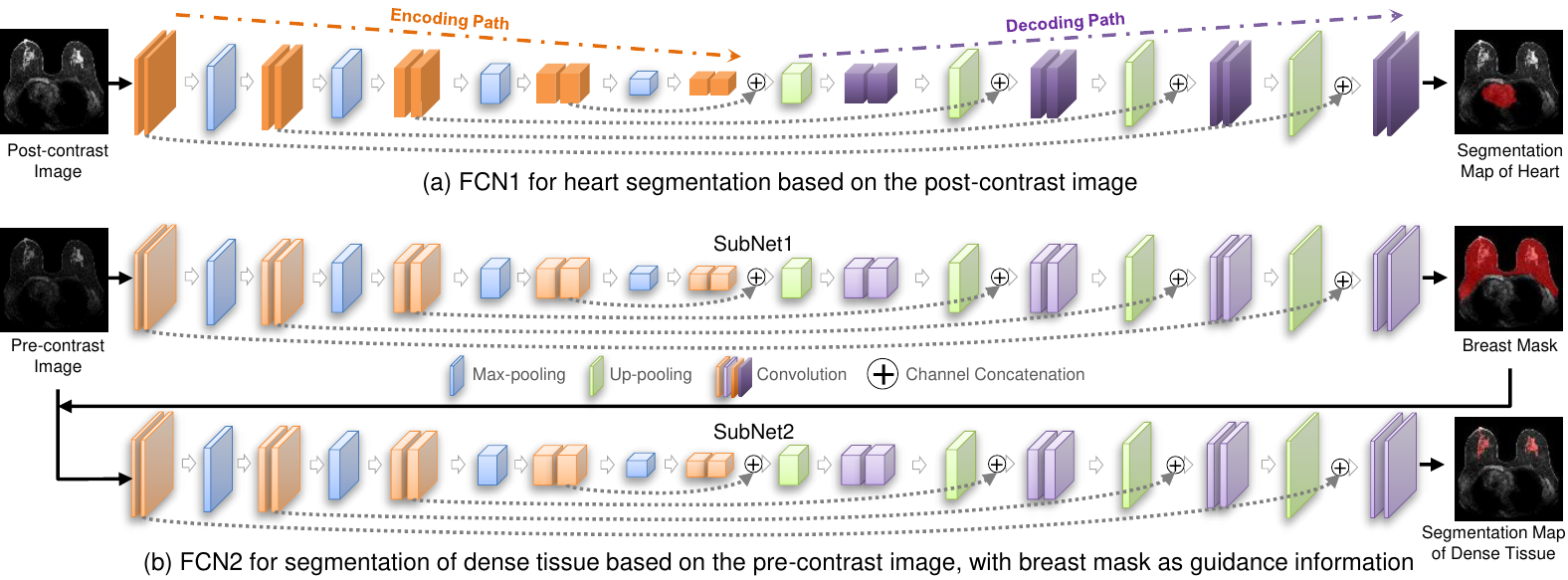}
	\caption{Illustration of the proposed FCN model for the segmentation of three types of tissues (\ie, the heart, dense tissue, and fat tissue). There are two sub-networks, including (a) FCN1 for the segmentation of the heart, and (b) FCN2 for the segmentation of dense tissues.}
	\label{fig_FCN}
\end{figure*}
To this end, a piecewise linear normalization method based on tissue segmentation is proposed for image intensity correction in this study. Our assumption here is that the same tissue (\eg, fat or dense tissue) in different images would share a similar intensity range. Then, it is possible to normalize the DCE-MR images by treating different tissues (\ie, air, fat tissue, dense tissue, and heart) separately. Specifically, a deep-learning-based segmentation method is first developed to perform tissue segmentation accurately and efficiently. Then, using the segmentation results, a subject-specific piecewise linear mapping function is proposed to normalize the same type of tissues in different images into a consistent intensity range. Experimental results on 100 testing subjects with DCE-MRI data demonstrate the effectiveness of the proposed method.

\section{Materials and Methods}
\subsection{Study Population and Imaging Protocol}
In this study, we analyzed data for 460 subjects with DCE-MR images which were collected with the Institutional Review Board approval including a waiver of informed consent. These images were acquired in the axial plane by using a 1.5T or 3.0T scanner (GE Healthcare and Siemens) in the prone position. In our dataset, each subject has one pre-contrast and three post-contrast T1-weighted sequences acquired after the intravenous administration of contrast agent [gadopentetate dimeglumine ( Magnevist, BayerHealth Care, Berlin, Germany)] or [gadobenate dimeglumine (MultiHance, Bracco, Milan, Italy)] using a weight based protocol (0.2 ml/kg). In the experiments, the dataset is randomly divided into two subsets, including the training set containing 360 subjects and the testing set with the remaining 100 subjects. The detailed imaging parameters for all data are organized in Table~\ref{tab:subjects}.

\begin{table}[t]
	\footnotesize
	\centering\renewcommand{\arraystretch}{1.4}
	\caption{Statistical information of imaging parameters for training and test datasets.}
	\scriptsize
	\begin{center}
		\begin{tabular*}{0.4\paperwidth}{lccccccc}
			\toprule
			& \multicolumn{2}{c}{Magnetic Field} & \multicolumn{2}{c}{TE}   & \multicolumn{2}{c}{TR}\\
			
			&$1.5$T &$3.0$T & $>2.0$ms & $<2.0$ms &$>4.5$ms&$<4.5$ms \\\hline
			Training Set &149	&211   &265 &95&265&95    \\
			Test Set	&48		&52	&79 & 21 & 79 &21 \\
			\bottomrule
		\end{tabular*}
	\end{center}
	\label{tab:subjects}
\end{table}

Our normalization algorithm, illustrated in Fig.~\ref{outline} consists of two steps: (1) different regions of the image, corresponding to different types of tissue/material are segmented using a deep learning algorithm, and (2) the pixel values of the image are adjusted using a piecewise linear transformation in order to bring the pixel value to a predefined range. In this study, we use the pre-contrast image and the first post-contrast image of each subject to obtain the parameters for image normalization.


\subsection{Step 1: Deep-learning-based Tissue Segmentation}
The breast MR images in this study contain four types of background, tissues or organs, including 1) air, 2) fat tissue, 3) dense tissue, and 4) the heart. For the air, we simply select the weakest $5\%$ (regarding intensity) as the air tissue. Recently, fully convolutional neural networks (FCNs) have been broadly used for anatomical tissue segmentation and organ detection~\cite{zhang2017detecting,cao2018deformable,zhang2017joint,pereira2016brain,lian2018multi,liu2018landmark,xu2017gland,bi2017dermoscopic,yap2017automated}. Inspired by the success of FCNs in medical image analysis, two deep learning models (\ie, two FCNs with the U-Net architecture~\cite{ronneberger2015u}) are proposed to accurately and efficiently segment the remaining three types of tissues (\ie, fat tissue, dense tissue, and the heart) from each MR image, with illustration shown Fig.~\ref{fig_FCN}.

In the \emph{first} FCN (FCN1 in Fig.~\ref{fig_FCN}~(a)), we aim to segment the heart from post-contrast images, because the intensity contrast between the heart and the other organs in the post-contrast image is much higher than that in the pre-contrast image. Specifically, FCN1 employs the U-Net architecture to learn a non-linear mapping from the input post-contrast image to the segmentation map for the heart. There are two major components in FCN1, including an encoding path and a decoding path. In the encoding path, every step contains one $3\times3\times3$ convolutions, followed by a batch normalization layer, a rectified linear unit (ReLU), and a $3\times3\times3$ max pooling operation with stride $2$ for down-sampling. In the decoding path, each step consists of a $3\times3\times3$ up-convolution, followed by concatenation with the corresponding feature map from the encoding path and one $3\times3\times3$ convolution (each followed by a ReLU). A Dice loss function~\cite{zhang2017detecting} is used in the last layer of FCN1 for generating a segmentation map for the heart. Hence, this network can not only grasp a large image area using small kernel sizes but also still keeps high segmentation accuracy, due to the encoding and decoding paths.

In the \emph{second} FCN (\ie, FCN2 in Fig.~\ref{fig_FCN}~(b)), we first segment the breast from the pre-contrast image in the first stage, and then segment the dense and fat tissues based on the breast region in the second stage. Since breast tissues (\ie, dense tissue and fat tissue) appear in the breast region, we would like to generate a region of interest (ROI) that can exclude confounding regions (\eg, air and heart), by using a breast mask as the ROI to remove the most enhanced organs. Specifically, there are two cascaded sub-networks (\ie, SubNet1 and SubNet2) in FCN2. The first stage (\ie, SubNet1) aims to generate a breast mask from the whole input image, while the second stage (\ie, SubNet2) is used to segment the dense tissue from the fat tissue in the breast region. The input of SubNet1 is a pre-contrast image, while the output is a breast mask. In SubNet2, the input is the masked image (using segmented breast region generated by SubNet1 as a mask to select the region of interest from the pre-contrast image), and the output is the segmented dense tissue. Here, SubNet1 and SubNet1 share the similar architecture as FCN1, while the difference is that the activation function of the last layer in both SubNet1 and SubNet2 is a sigmoid function to normalize the output into $[0,1]$. In particular, based on the segmented dense tissue, the fat tissue can be generated by a subtraction between the breast masked image and the segmented map of dense tissue. After obtaining the dense tissue, we can obtain the fat tissue by excluding the dense tissue from the breast mask. Therefore, using the proposed FCN1 and FCN2 models, three types of tissues (\ie, the heart, dense tissue, and fat tissue) can be segmented from the input pre-contrast and post-contrast MR images for each subject.

\subsection{Step 2: Piecewise Linear Mapping}

In this work, we assume that the same type of tissues (\eg, fibroglandular tissue) or material from different subjects should have similar image intensity, and the intensities of different tissue types would be different. Accordingly, a piecewise linear mapping method is proposed to normalize the image intensity of different subjects into a common space, based on the segmentation results for different tissues generated by the above-mentioned FCN model.

First, we defined a typical value for each tissue/material by identifying an \textit{archetype} subject in our training data in the following way. In the \emph{training} stage, the common tissue intensity for each type of tissue is first computed from the training MR images. We select the median image from all training MR images using the average rank of four median tissue values. (see next paragraph on the details of calculating the median value for the heart). Specifically, for the given $N$ training images, we can obtain a vector $\m^t=\left[m^t_1, m^t_2, \cdots, m^t_N\right]$ ($t=1,\cdots,4$) for four types of tissues. For each type of tissue, we can get a rank of images using the median tissue value. By averaging four ranks, we select the median subject from all training data. Here, the common value tissue intensity, denoted as $M^t$, is then computed as the median intensity from the median subject in the $t$-th tissue. That is, there are four intensity values ${\left\lbrace M^t \right\rbrace}^4_{t=1}$ in the common intensity space, with each one denoting the common value for each tissue computed based on the training data.

In the \emph{testing} stage, given a new subject, pixel values of each tissue type are transformed to take values of the archetype patient. Specifically, the proposed FCN model is first applied to its pre-contrast and post-contrast images to generate the segmentation of four types of tissue/material. With the segmented tissues, we would like to construct a piecewise linear mapping between the testing image and the common intensity space, by treating each tissue individually. That is, the median value ${\left\lbrace V^t \right\rbrace}^4_{t=1}$ for each tissue in the testing image is normalized to its corresponding median value in the common intensity space (the archetype patient). Therefore, the control points for the piecewise linear mapping are ${\left\lbrace (V^t, M^t)\right\rbrace}^4_{t=1}$. Then all values in between of the control points in the test image are interpolated linearly between the two control points. For intensity values that are larger than the median intensity of the heart (\eg, the highest value), we simply extend the mapping trend between dense tissue and heart to get the slope of linear mapping. For example, as the curve shown in Fig. 0, the x-axis defines the intensity space of testing image and the y-axis defines the common intensity space of all training images. These points (\ie, ${\left\lbrace (V^t, M^t)\right\rbrace}^4_{t=1}$ ) means the median values of air, fat tissue, dense tissue, and heart.

Note that, the values for air, fat, and dense tissue (\ie, ${\left\lbrace V^t \right\rbrace}^3_{t=1}$ or ${\left\lbrace M^t \right\rbrace}^3_{t=1}$) are calculated from pre-contrast image, and the value for the heart (i.e., $V^4 $ or $M^4$) is calculated from the first post-contrast image. It is also worth noting that we employ the $10\%$ maximum tissue value for heart instead of using median value, because of the inhomogeneity of heart. For the other tissues, we employ the median tissue values. Finally, the same mapping function is applied to all pre- and post-contrast images.
\section{Evaluation}
The goal of the proposed normalization algorithm is transforming the images to the same framework where the pixel values are directly comparable. Therefore, the expectation is that that following the normalization, the average pixel values of different types of tissue are similar regardless of which scanner parameters were used. Furthermore, it is expected that following normalization, algorithmic features (computer-aided diagnosis features/radiomics features) do not systematically differ between sets of images acquired using different scanning parameters. Finally, we expect that image feature extracted from a set of normalized images will generally be as useful or more useful for prediction of various quantities (e.g. tumor genomic or patient outcomes) than the same features extracted from images that were not normalized. Following these expectations, we will evaluate our algorithm in the following steps:

\begin{itemize}
	\item Visually evaluate the intermediate step of deep learning-based segmentation.
	\item Evaluate whether the values for different tissue types (before and after normalization) depend on scanner parameters.
	\item Evaluate whether algorithmically extracted features (before and after normalization) depend on scanner parameters.
	\item Evaluate whether the features extracted from normalized images have a similar or generally higher prognostic value for prediction of tumor molecular subtypes.
	\item Evaluate time efficiency of our software.
\end{itemize}

\begin{figure}
	\setlength{\abovecaptionskip}{-0pt}
	\setlength{\belowcaptionskip}{-2pt}
	\setlength\abovedisplayskip{-2pt}
	\setlength\belowdisplayskip{-2pt}	
	\centering
	\includegraphics[width=0.5\textwidth]{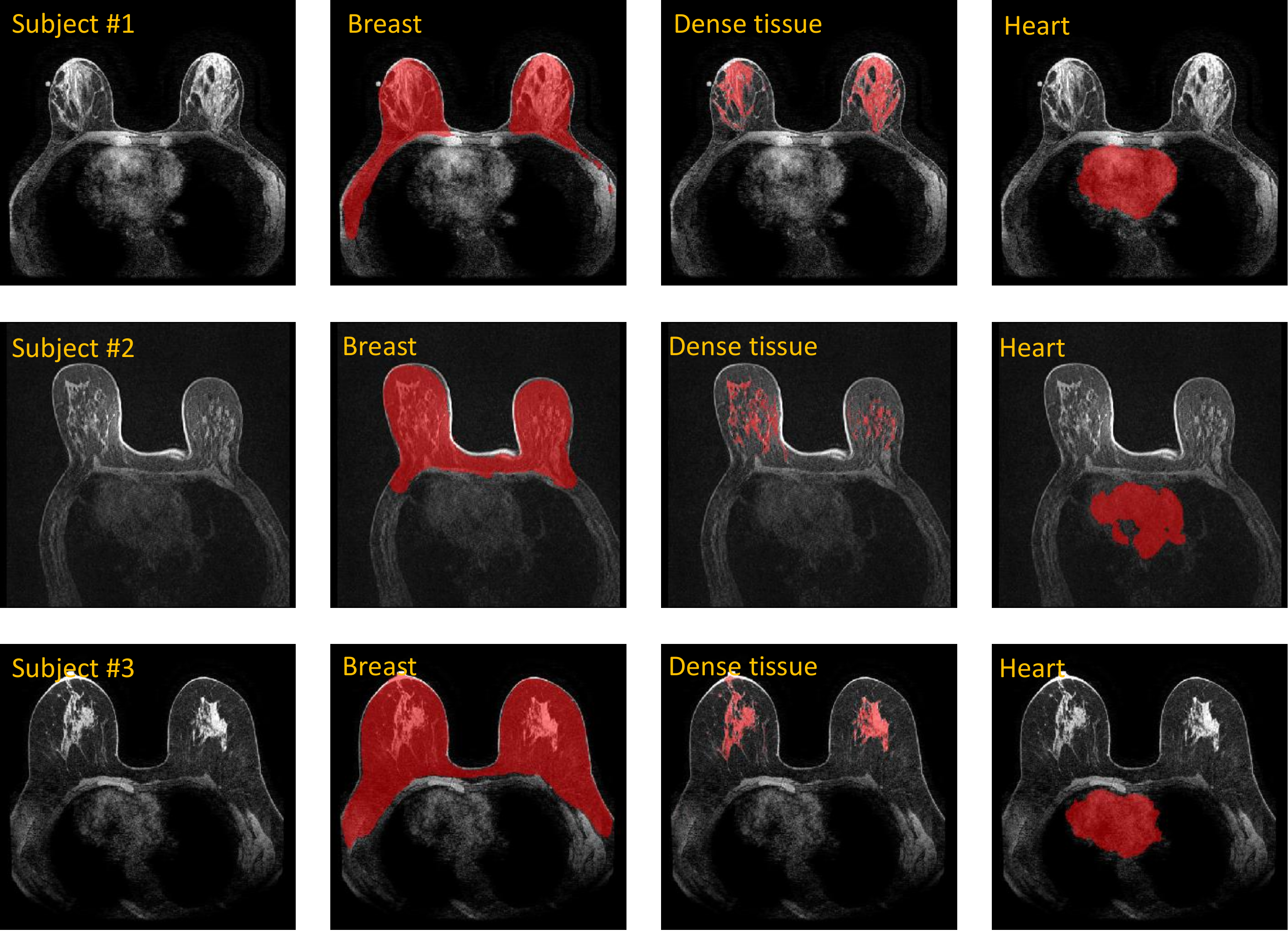}
	\caption{Segmentation results of the breast (\ie, fat tissue + dense tissue), dense tissue, and the heart for three typical subjects, achieved by our method, with each row corresponding to a particular subject.}
	\label{segmentation}
\end{figure}

To validate the effectiveness of our image normalization method in regard to feature extraction and radiogenomics analysis (points 3 and 4), we extracted 15 features (denoted as F1-F15) from the original and the normalized MR images (using one pre-contrast image and three post-contrast images for one subject) achieved by our method. The features represent different aspects of the tumor and normal breast parenchyma in DCE-MRI and depend on the intensity of the sequences. One of the features extracted is the algorithmically calculated major axis length~\cite{grimm2015computational} of the tumor (F9). This feature is based only on the semi-automatic tumor mask and can potentially indicate the effect of normalization on the segmentation of the tumor. Next, mean and standard deviation of fat tissue, healthy dense tissue, and tumor were extracted from the first post-contrast sequence resulting in the total of 6 features (F10-F15). In addition to these features, enhancement properties of the tumor were calculated based on voxel-wise SER (signal enhancement ratio) based mean~\cite{arasu2011can} (F2) and voxel-wise washin rate based mean (F4) and standard deviation (F5)~\cite{arasu2011can,wang2015identifying}. These features are dependent on the intensity values of multiple sequences (pre-contrast and multiple post-contrast sequences). Next, one tumor enhancement texture based feature is calculated based on a dynamic histogram of oriented gradients ~\cite{wan2016radio} (F8) as it considers the spatial relationship of tumor voxels. Also, healthy dense tissue enhancement-based features were calculated through the mean of voxel-wise washin rate map (F3) of tissue and mean and standard deviation of the voxel-wise SER of tissue~\cite{arasu2011can,wang2015identifying} (F1 and F6). The enhancement texture of tissue is calculated using the entropy of the voxel-wise percentage enhancement of the tissue~\cite{arasu2011can,wang2015identifying,haralick1973textural} (F7).

Finally, once the features were extracted based on the original images and the normalized images, we compared these sets of features for the task of predicting tumor molecular subtypes. This task is of particular interest in recent years and is a part of the emerging discipline of radiogenomics. Specifically, we attempted to distinguish the Luminal A subtype from the remaining subtypes including luminal B, human epidermal growth factor receptor 2 (HER2) enriched, and triple negative (3,4).


\subsection{Tissue Segmentation}

We illustrate the segmentation results of the breast (i.e., fat tissue + dense tissue), dense tissue, and the heart for typical testing subjects in Fig.~\ref{segmentation}. Since the ground-truth segmentation of test subjects is unknown in this study and the segmentation is only a mean toward the normalization goal, we do not offer a quantitative assessment of this step. This figure suggests that our proposed deep learning method can perform very good segmentation of three types of tissues based on both pre-contrast and post-contrast MR images. We released our trained model for tissue segmentation as well as the code for testing online (https://github.com/MaciejMazurowski/breast-mri-normalization). It is worth noting that the proposed FCN-based method requires less than 30 seconds to finish segmentation of three types of tissues, suggesting the efficacy of our method. Finally, please note that since the segmentation is used to establish a typical value of a certain type of tissue rather than determining the exact boundaries, some imperfection in segmentation is acceptable.



\begin{figure*}[tp]
\setlength{\belowcaptionskip}{-2pt}
\setlength\abovedisplayskip{-2pt}
\setlength\belowdisplayskip{-2pt}
	\centering
	\includegraphics[width=\textwidth]{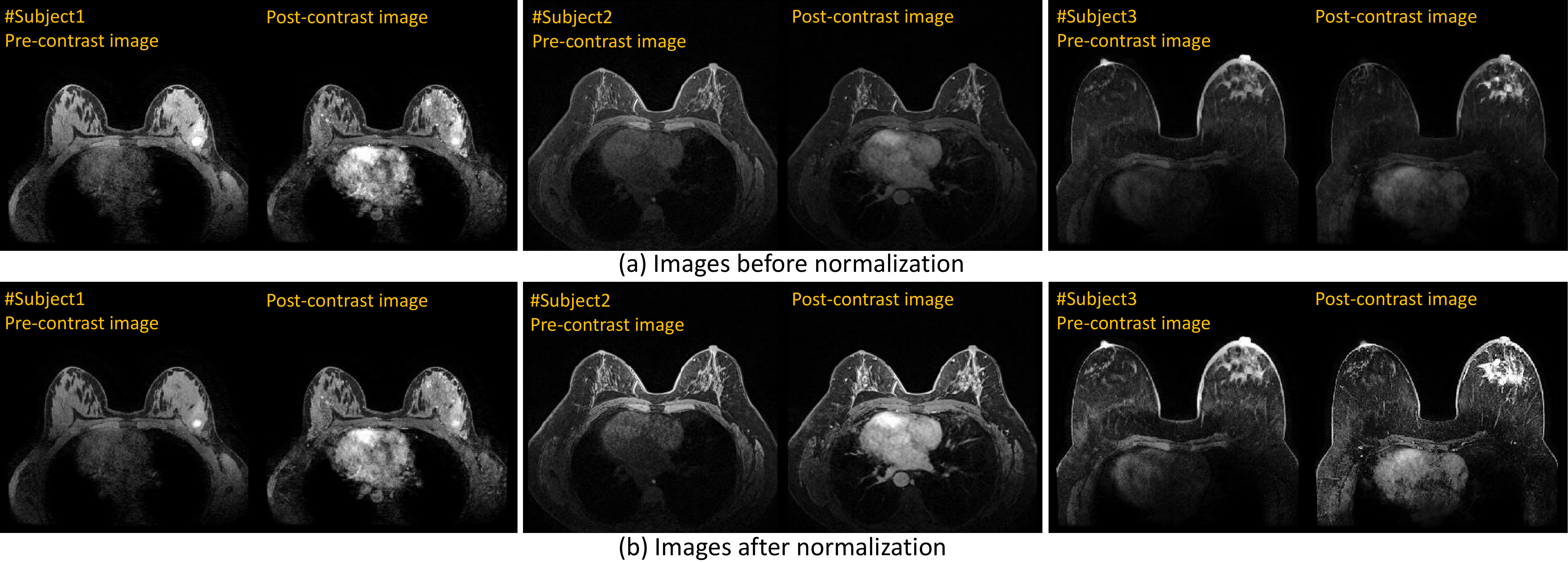}
	\caption{Qualitative illustration of typical images before and after normalization. (a) Images before normalization. (b) Images after normalization.}
	\label{qualitativeshow_norm}
\end{figure*}

\begin{figure*}[tbp]
\setlength{\belowcaptionskip}{-2pt}
\setlength\abovedisplayskip{-2pt}
\setlength\belowdisplayskip{-2pt}
	\centering
	\begin{subfigure}[b]{1\textwidth} 
		\includegraphics[width=1\textwidth]{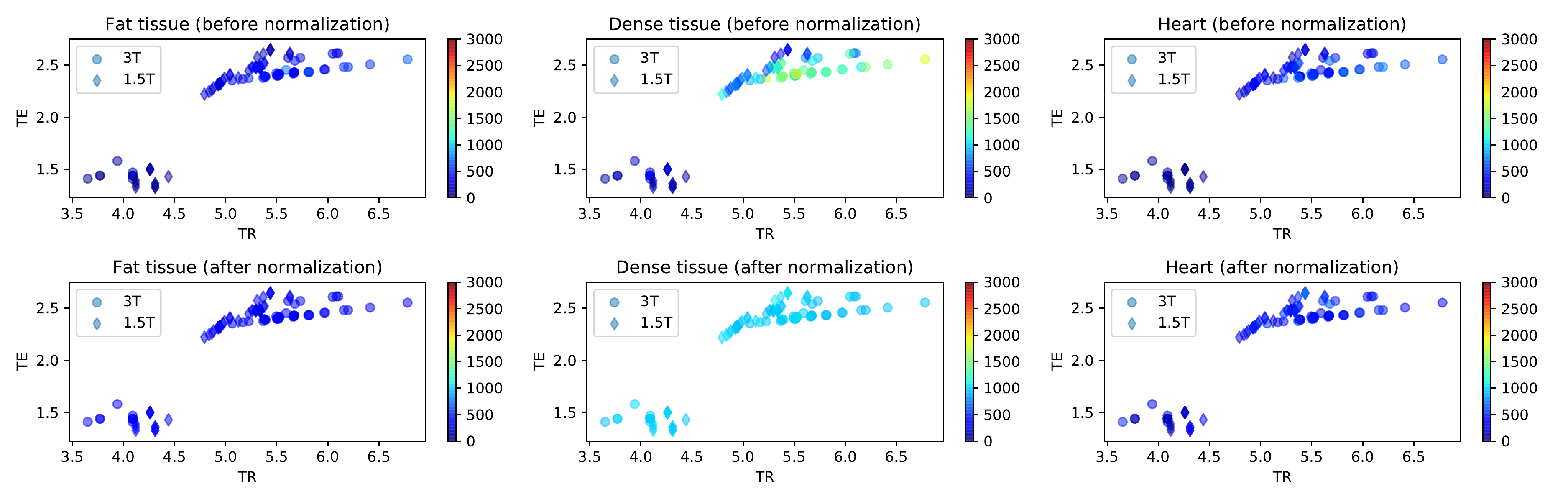}
		\label{figure_pre}
		\caption{Pre-contrast image}
		
	\end{subfigure}
	
	\begin{subfigure}[b]{1\textwidth} 
		\includegraphics[width=1\textwidth]{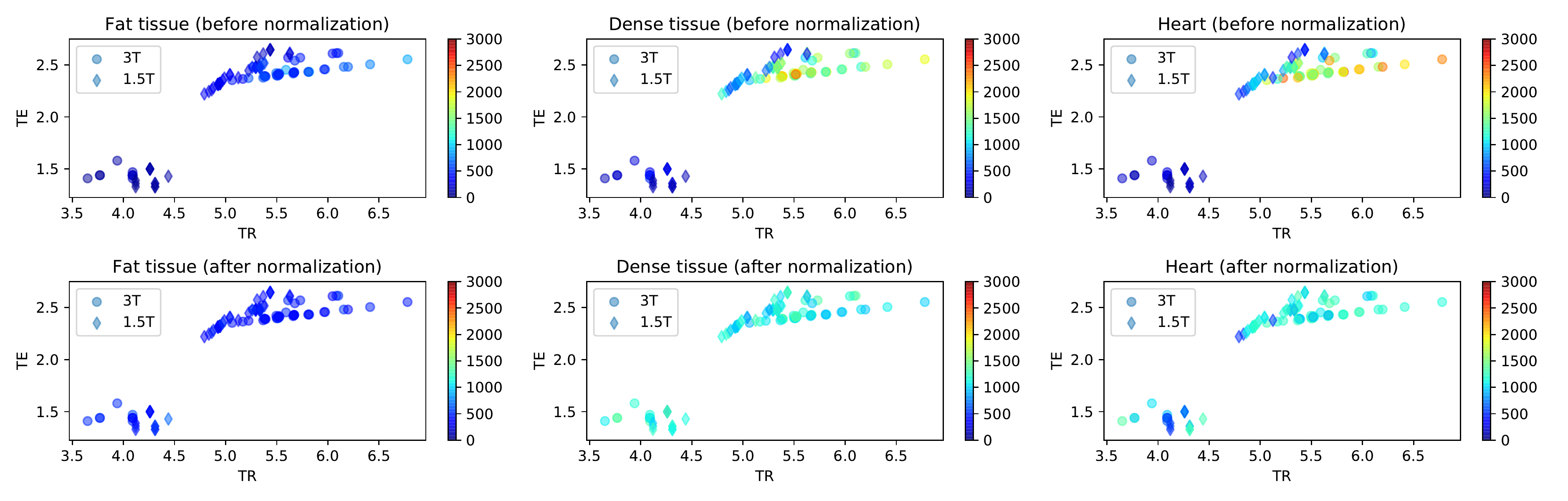}
		\label{figure_post}
		\caption{Post-contrast image}
	\end{subfigure}
	\caption{Mean intensity values of different tissues in DCE-MRI. (a) Pre-contrast image. (b) Post-contrast image. } 
	\label{figure_prepost}
\end{figure*}

\begin{figure*}
	\setlength{\belowcaptionskip}{-2pt}
	\setlength\abovedisplayskip{-2pt}
	\setlength\belowdisplayskip{-2pt}
	\centering
	\begin{subfigure}[b]{0.9\textwidth} 
		\includegraphics[width=\textwidth]{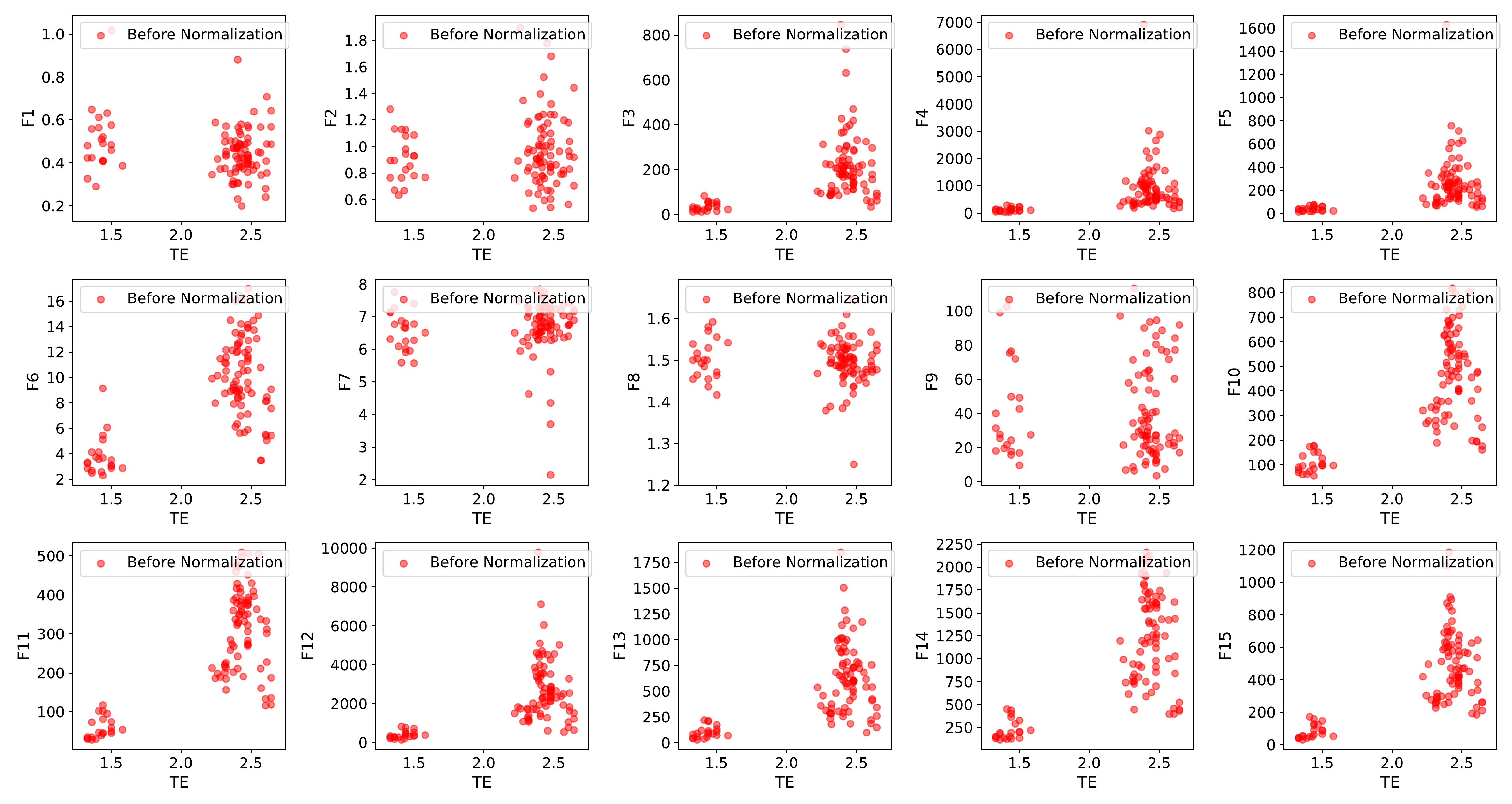}
		\label{feature_te_before}
		\caption{TE}
		
	\end{subfigure}
	
	\begin{subfigure}[b]{0.9\textwidth} 
		\includegraphics[width=\textwidth]{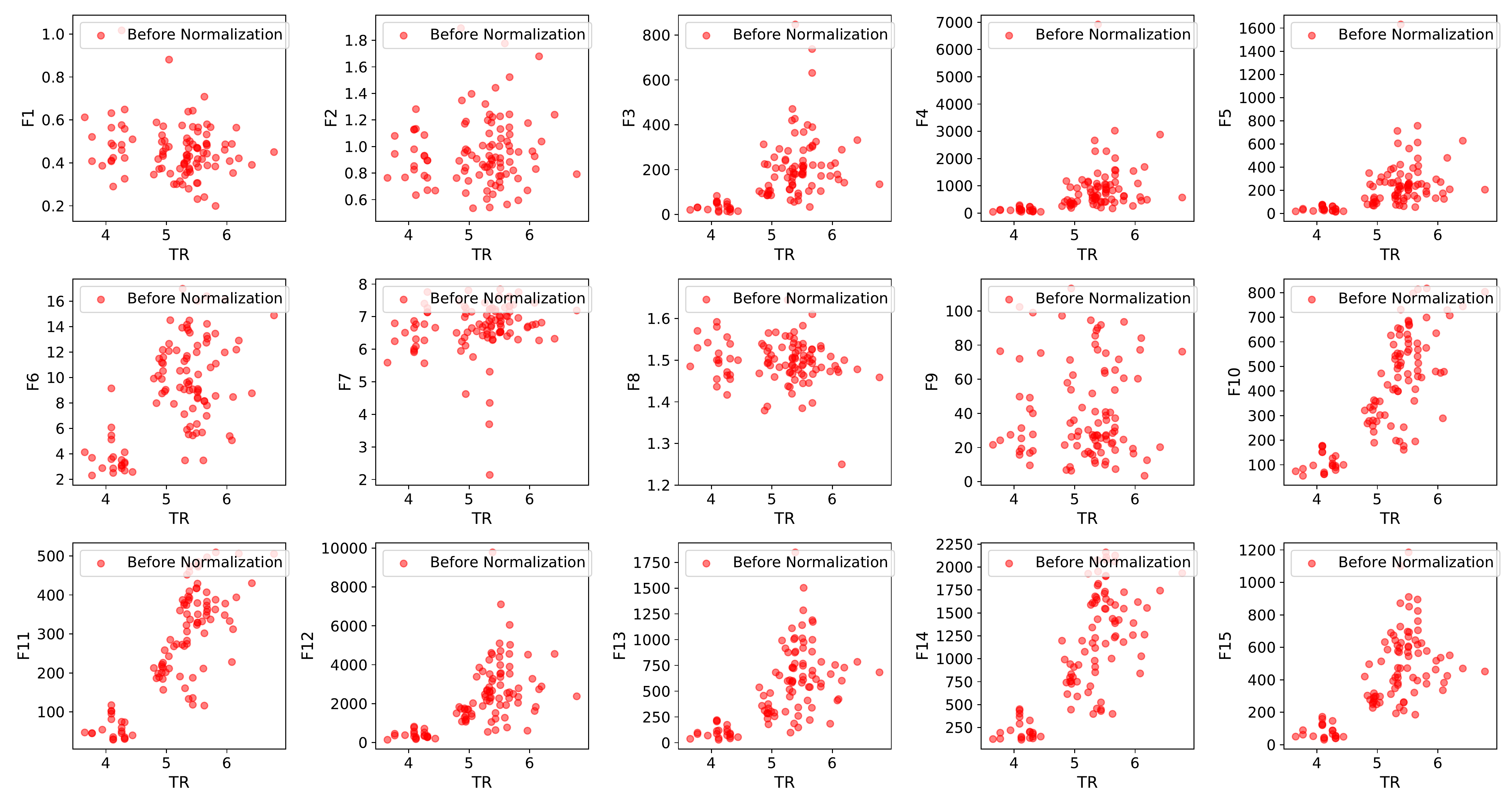}
		\caption{TR}
		\label{feature_tr_before}
	\end{subfigure}
	\caption{Feature distribution with different imaging parameters for the original images. (a) TE and (b) TR.} 
	\label{feature_before}
\end{figure*}

\subsection{Standardization of Pixel Values for Different Types of Tissue}
Figure~\ref{qualitativeshow_norm} shows the original images with different imaging parameters (TR, TE, and magnet strength) as well as normalized images. After intensity normalization using our method, these images share very similar intensity-range for each specific tissue. 

Figure~\ref{figure_prepost} illustrates average values (represented by color) as they relate to different scanner parameters including TE and TR (on the axes of the plot) and magnet strength (marker type). A strong relationship is seen between the mean pixel values and the scanner parameters particularly for the normal breast parenchyma which is crucial in the analysis of breast cancer images. This relationship is clear for both pre-contrast and post-contrast images. Following the normalization, no clear relationship between the scanner parameters and pixel values can be seen which is the desired effect.


\subsection{Impact of Normalization on Feature Values}
\begin{figure*}
\setlength{\belowcaptionskip}{-2pt}
\setlength\abovedisplayskip{-2pt}
\setlength\belowdisplayskip{-2pt}
	\centering
	\begin{subfigure}[b]{0.9\textwidth} 
		\includegraphics[width=\textwidth]{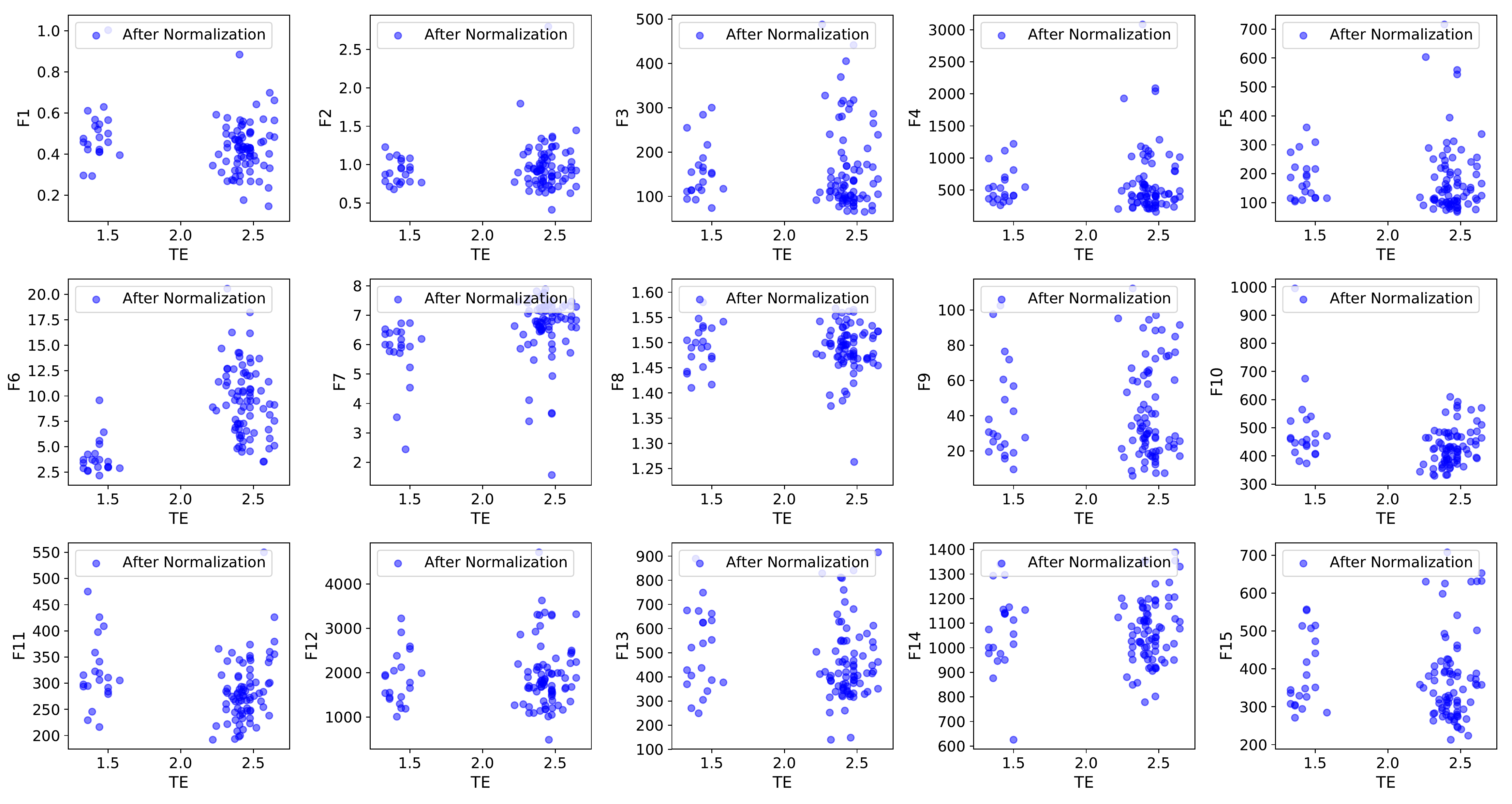}
		\caption{TE}
		\label{feature_te_after}
		
	\end{subfigure}
	
	\begin{subfigure}[b]{0.9\textwidth} 
		\includegraphics[width=\textwidth]{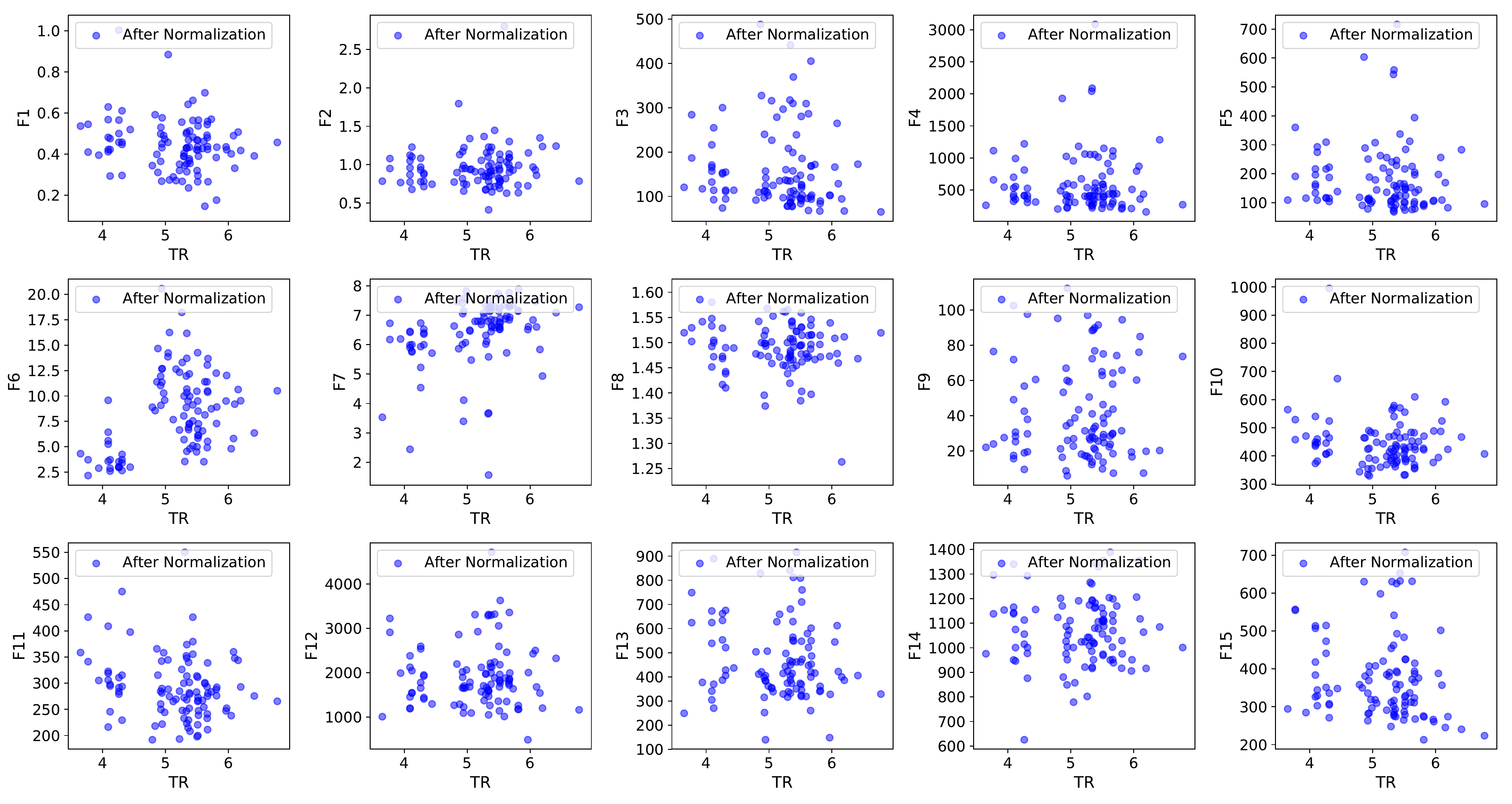}
		\caption{TR}
		\label{feature_tr_after}
	\end{subfigure}
	\caption{Feature distribution with different imaging parameters for the images after normalization using the proposed method.  (a) TE and (b) TR. } 
	\label{feature_after}
\end{figure*}

As shown in Fig.~\ref{feature_before}, five features (\ie, F1, F2, F7, F8, and F9) extracted from the \emph{original images} are not affected by different values of TE and TR (\ie, sharing the similar data distribution), while the rest features are highly dependent on the values of TE and TR (\ie, having different data distribution). As shown in Fig.~\ref{feature_after}, all features extracted from our \emph{normalized images} are not affected by the values of TE and TR, except for F6 (\ie, the standard deviation of SER map in the dense tissue region). 


\begin{figure*}
	\centering
	\includegraphics[width=0.8\textwidth]{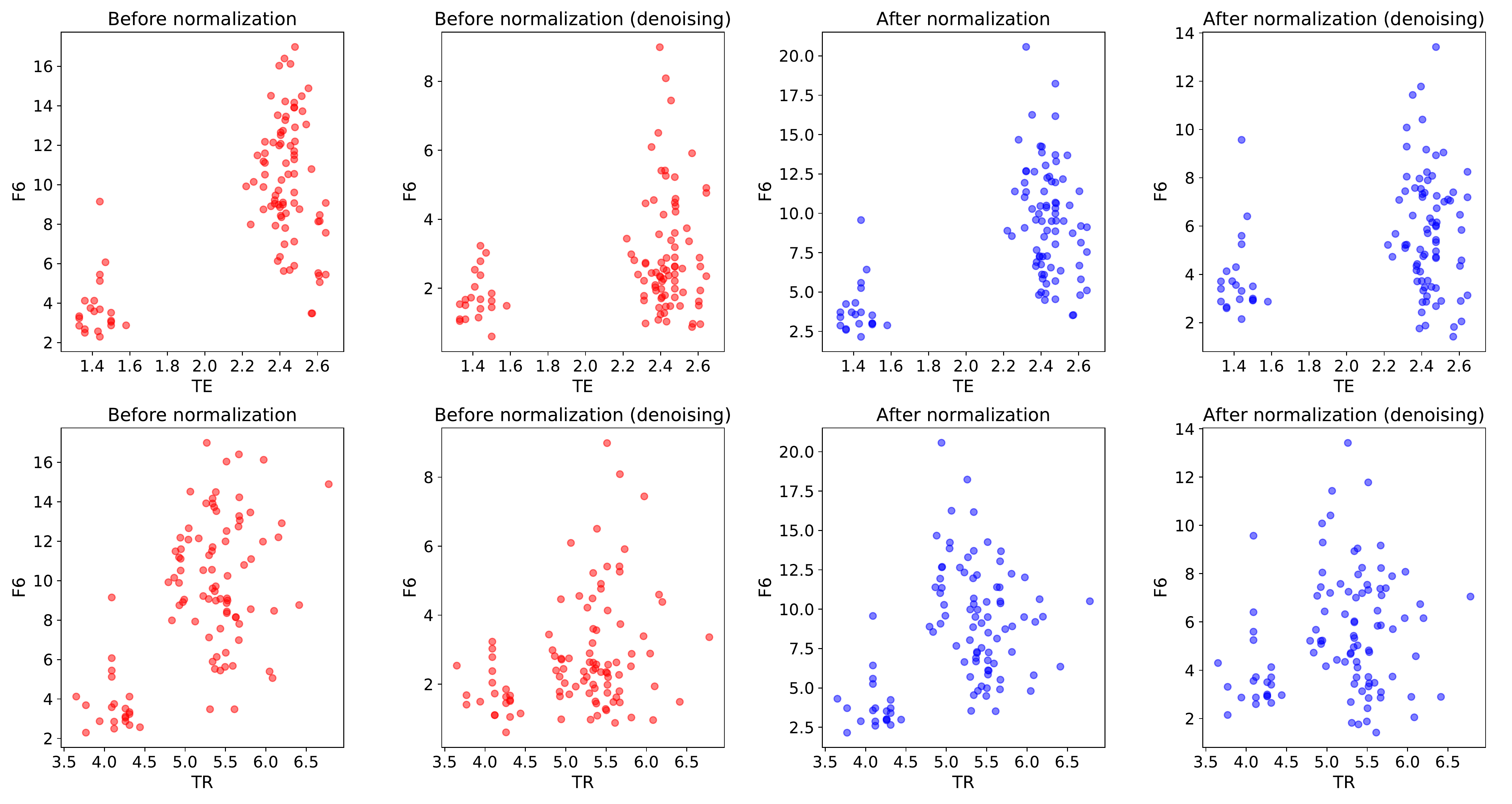}
	\caption{Feature distribution of F6 concerning TE and TR, using the denoising strategy. The first row is for the parameter of TE and the second row is for the parameter of TR. }
	\label{denoising}
\end{figure*}

\subsection{Impact of Normalization on Radiogenomic Analysis}
\begin{figure*}[tp]
	\setlength{\abovecaptionskip}{-2pt}
	\setlength{\belowcaptionskip}{-10pt}
	\setlength\abovedisplayskip{-10pt}
	\setlength\belowdisplayskip{-10pt}
	\centering
	\includegraphics[width=1\textwidth]{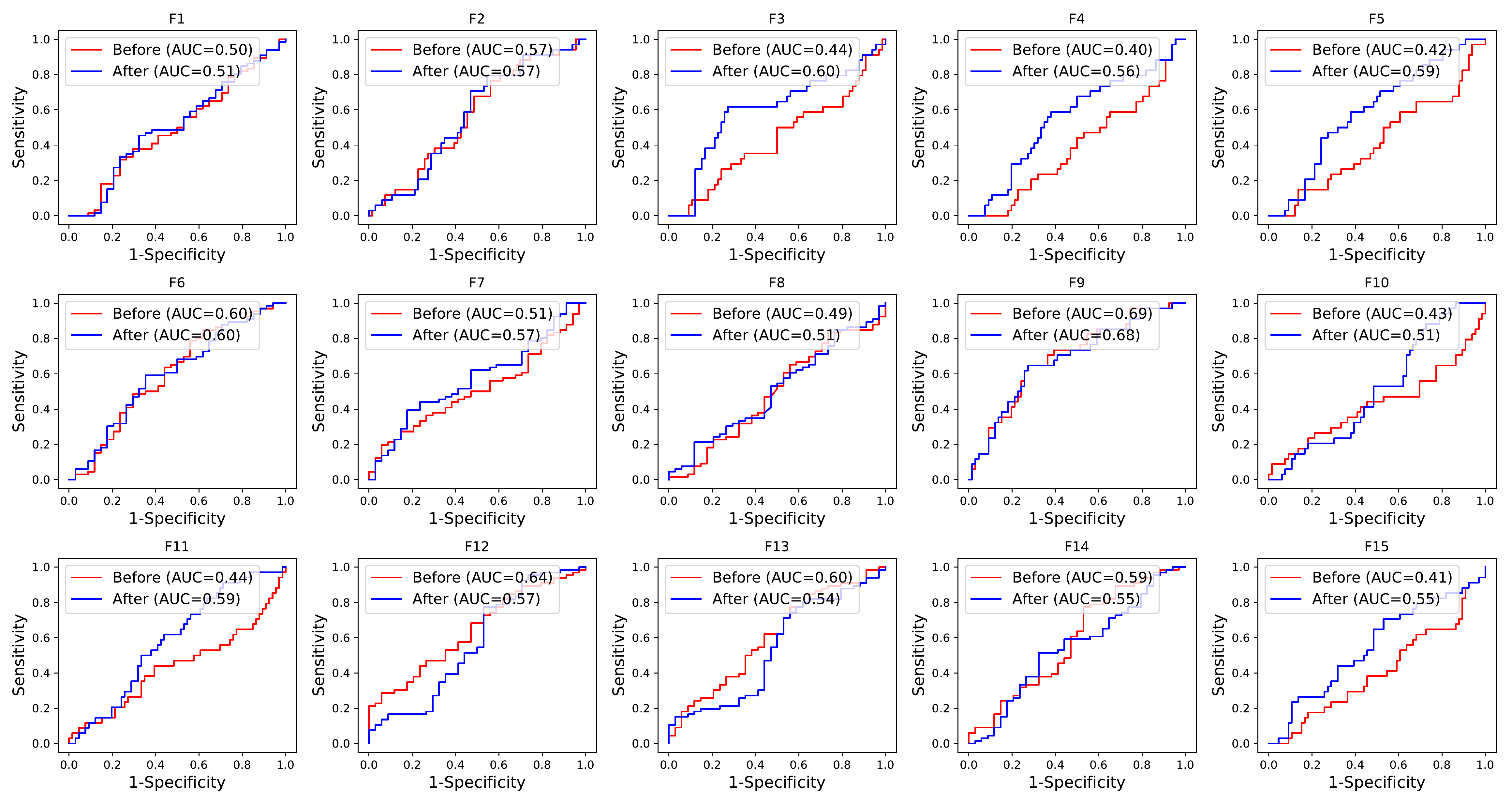}
	\caption{ROCs based on fifteen features (\ie, F1-F15) in identifying luminal A from four types of tumor subtypes, including 1) luminal A, 2) luminal B, 3) human epidermal growth factor receptor 2 enriched, and 4) basal-like (3,4).}
	\label{figure_ROC}
\end{figure*}

We plot the Receiver Operating Characteristic (ROC) curves as well as corresponding Area Under the ROC Curve (AUC) based on each feature in Fig.~\ref{figure_ROC} separately. We observe that, the AUC values increase, sometimes notably, using the majority (10/15). We did not observe a change in two features and we observed a slight decrease in three features. These results demonstrate that our MR image intensity normalization method can potentially improve the discriminative capability of extracted features for radiogenomics analysis, implying the potential of the proposed method in real-world applications.


\subsection{Computational Efficiency}
An important advantage of our algorithm is that it is relatively fast. In Table~\ref{tab_time}, we report the computational cost for breast segmentation, heart segmentation, dense tissue segmentation, and intensity mapping respectively, using one Nvidia GPU (GTX 1080) with 8 GB GPU RAM. From the table, we can notice that our method requires only less than 30 seconds to complete the entire normalization process. Note that, our speed is also limited by the memory of GPU, since we cannot predict the entire segmentation map for each image. We have to split the images into many subimages and predict them separately, and then reconstructing them into one segmentation map. This is time-consuming since the convolution calculation is repeated multiple times for many voxels. Therefore, the speed of our normalization can be further improved using GPU with large memories.

\subsection{Limitations: Normalizing Average Pixel Values VS. Changing Noise Characteristics}
We observed that for a one analyzed feature, namely standard deviation of SER, our normalization algorithm did not result in a significant alignment of the features values for different scanner parameters as observed for all other analyzed features. The likely reason is that the SER map is more sensitive to data noise. Generally, the mean value of SER map can potentially remove the effect of noise, but the standard deviation can be easily affected by the noise. To further investigate this issue, we apply a de-noising strategy using a median filter to the high TE/TR images, and find the resulting features (\ie, standard deviation of the voxel-wise SER of tissue) with different TE/TR become closer in their distribution, as shown in Fig.~\ref{denoising}. These results imply that the proposed normalization method can remove the effect of the overall intensity variation for feature extraction, and could not mitigate the effect of image noise in this specific case.

Additionally, In Fig.~\ref{mapping}, we show the mapping functions for 25 cases randomly selected from the testing set. Fig.~\ref{mapping} demonstrates that different slopes are used for normalizing different tissues, to ensure that the same type of tissue among different images would have the similar intensity range.
\begin{figure}[tp]
	\centering
	\includegraphics[width=0.5\textwidth]{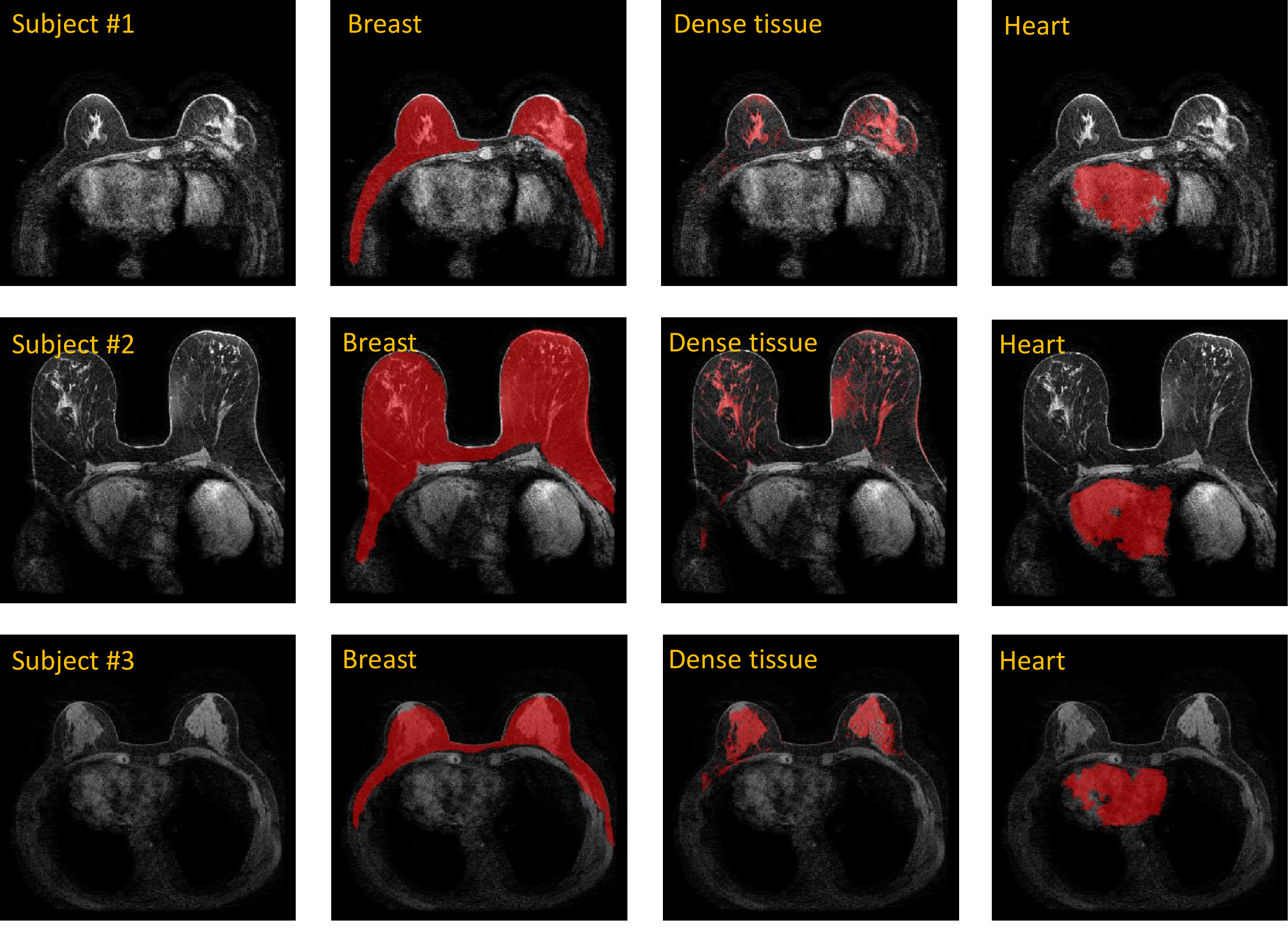}
	\caption{Inaccurate training mask of heart, breast, and dense tissue for three typical subjects.}
	\label{qualitativeshow_training}
\end{figure}

\begin{figure*}[htbp]
	\setlength{\abovecaptionskip}{-2pt}
	\setlength{\belowcaptionskip}{-10pt}
	\setlength\abovedisplayskip{-10pt}
	\setlength\belowdisplayskip{-10pt}
	\centering
	\includegraphics[width=0.98\textwidth]{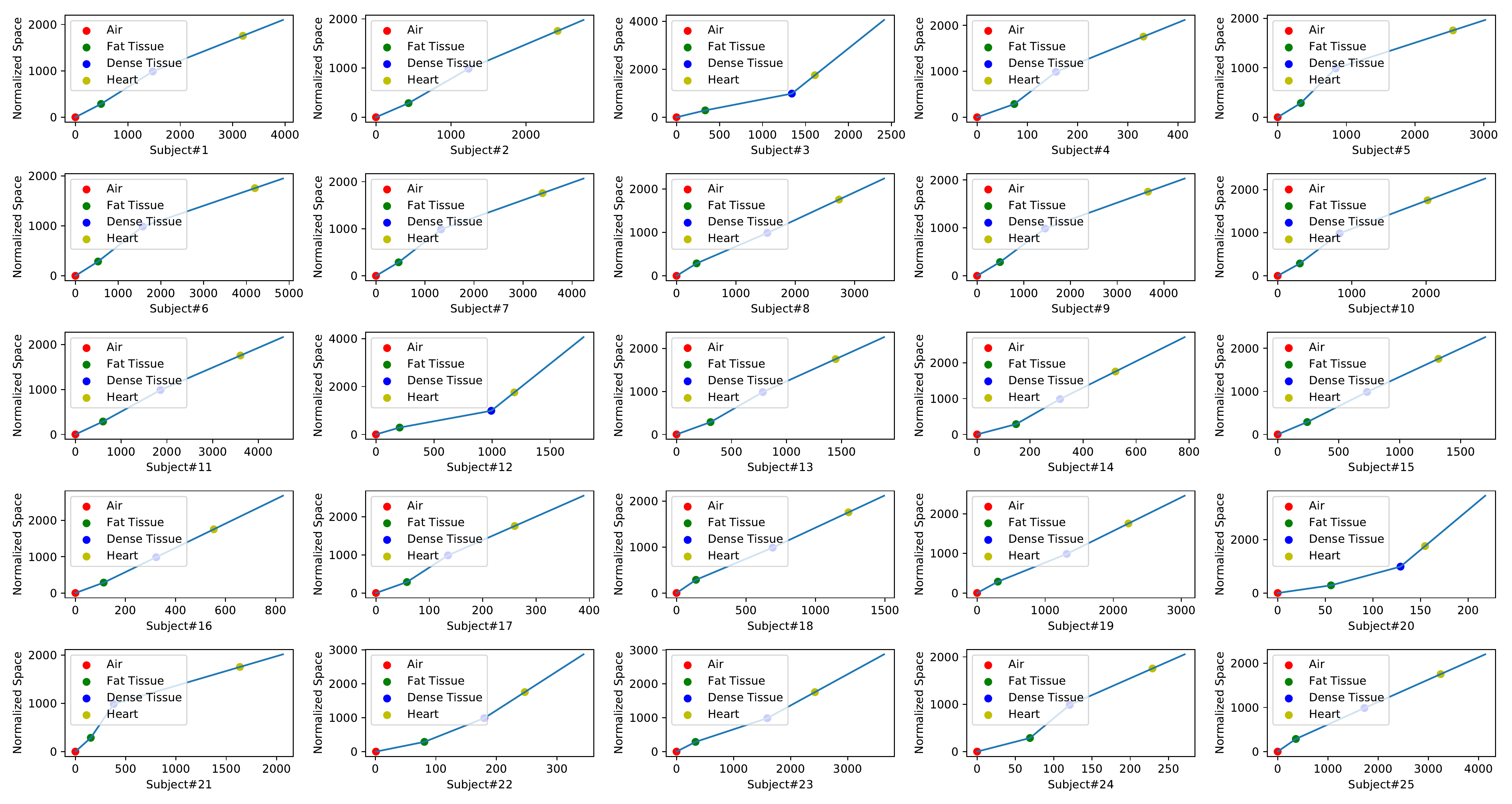}
	\caption{The proposed piecewise linear mapping for $25$ cases in the testing set. The x axis defines the intensity space of testing image and the y axis defines the common intensity space of all training images.}
	\label{mapping}
\end{figure*}

\section{Discussion}
In this paper, we proposed an intensity normalization algorithm for breast DCE-MRI volumes which uses a deep learning-based segmentation of different parts of the image and a piecewise linear mapping. The algorithm is based on an assumption that the same type of tissue should share a similar range of intensity values in DCE-MRI in different patients. We have thoroughly tested our algorithm and showed that it successfully normalizes the intensity of DCE-MR images. We made our software publicly available for others to apply in their analyses.

Our algorithm has multiple advantages. It does not require any physics information which might not be available in some datasets. Also, our method is computationally efficient. The piecewise linear mapping also guarantees monotonic continuity in the intensity space. 

Our algorithm and study have some limitations. First, by design, the algorithm focuses on normalization of overall pixel intensities rather than noise characteristics of the images which might differ between different scanners. This was illustrated with the difficulty with F 6 (\ie, the standard deviation of the voxel-wise SER of tissue feature). Please note, however, that the algorithm was able to normalize the images in a way that addressed the disparity for most of the features related to nose such as texture features. A further limitation of our algorithm is that our algorithm used only axial images (the most typical view for breast MRI). Finally, while we used a broad set of scanners, the list was not exhaustive. This limits the training of our algorithm as well as the evaluation. 

We performed a cursory evaluation on a few selected cases from a publicly available dataset with generally acceptable results but with some issues, potentially dues to different imaging views (not axial), very different resolutions, and older scanners. We encourage the readers of this paper to evaluate our algorithm on their datasets and share their results.

It is worth noting that we only employ very rough segmentation masks of the breast, dense tissue, and heart for training the model. These training masks are generated by very basic unsupervised segmentation method with a few manual corrections. An example of the training mask is shown in Fig.~\ref{qualitativeshow_training}, where we can observe an imperfect but generally accurate generated mask. As shown in Fig.~\ref{segmentation}, our predicted masks are even better than the quality of training masks. This phenomenon was previously observed with machine learning models because of statistical property~\cite{zhang2016detecting}. Since a highly accurate segmentation is not the goal of this study and the ground truth used was imperfect, we do not quantitatively assess the segmentation performance of our algorithm. However, visual examination of selected masks showed satisfactory results. Please note that our algorithm is highly tolerant of imperfections in the segmentation process as it only requires median values for different tissue/material types.

\begin{table}[h]
\renewcommand\arraystretch{1.4}
		\centering
		\caption{Computational costs of different steps in the proposed image normalization approach for one image of $512\times 512 \times 90$ with the spacing of $0.68\times 0.68 \times 2 mm^3$.}
		\footnotesize
		\begin{tabular*}{0.45\textwidth}{ccccc}
			\toprule
			&Breast  & Heart  & Dense tissue & Mapping    \\
			\hline			
			Computational time& 4.1 s & 4.9 s & 13.6 s &1.8 s  	\\

			\bottomrule
			
		\end{tabular*}
		\label{tab_time}
\end{table}

\section{Conclusion}
In this paper, we proposed an intensity normalization method for breast DCE-MRI. Specifically, we develop a piecewise linear mapping strategy to map the image intensity to a common space, using the segmentation results of fat, dense tissue, and heart. Our experimental results showed the mean value from one specific tissue has the similar intensity range among images scanned with different parameters. Also, we demonstrated that features extracted from our normalized images are more discriminate in molecular subtype classification, compared with features from the original images.

\bibliographystyle{IEEEtran}
\bibliography{mybib}

\begin{thebibliography}{10}
\providecommand{\url}[1]{#1}
\csname url@samestyle\endcsname
\providecommand{\newblock}{\relax}
\providecommand{\bibinfo}[2]{#2}
\providecommand{\BIBentrySTDinterwordspacing}{\spaceskip=0pt\relax}
\providecommand{\BIBentryALTinterwordstretchfactor}{4}
\providecommand{\BIBentryALTinterwordspacing}{\spaceskip=\fontdimen2\font plus
\BIBentryALTinterwordstretchfactor\fontdimen3\font minus
  \fontdimen4\font\relax}
\providecommand{\BIBforeignlanguage}[2]{{%
\expandafter\ifx\csname l@#1\endcsname\relax
\typeout{** WARNING: IEEEtran.bst: No hyphenation pattern has been}%
\typeout{** loaded for the language `#1'. Using the pattern for}%
\typeout{** the default language instead.}%
\else
\language=\csname l@#1\endcsname
\fi
#2}}
\providecommand{\BIBdecl}{\relax}
\BIBdecl

\bibitem{mahrooghy2015pharmacokinetic}
M.~Mahrooghy, A.~B. Ashraf, D.~Daye, E.~S. McDonald, M.~Rosen, C.~Mies,
  M.~Feldman, and D.~Kontos, ``Pharmacokinetic tumor heterogeneity as a
  prognostic biomarker for classifying breast cancer recurrence risk,''
  \emph{IEEE Transactions on Biomedical Engineering}, vol.~62, no.~6, pp.
  1585--1594, 2015.

\bibitem{mazurowski2015recurrence}
M.~A. Mazurowski, L.~J. Grimm, J.~Zhang, P.~K. Marcom, S.~C. Yoon, C.~Kim,
  S.~V. Ghate, and K.~S. Johnson, ``Recurrence-free survival in breast cancer
  is associated with {MRI} tumor enhancement dynamics quantified using computer
  algorithms,'' \emph{European Journal of Radiology}, vol.~84, no.~11, pp.
  2117--2122, 2015.

\bibitem{mazurowski2015radiogenomics}
M.~A. Mazurowski, ``Radiogenomics: {W}hat it is and why it is important,''
  \emph{Journal of the American College of Radiology}, vol.~12, no.~8, pp.
  862--866, 2015.

\bibitem{karahaliou2014assessing}
A.~Karahaliou, K.~Vassiou, N.~Arikidis, S.~Skiadopoulos, T.~Kanavou, and
  L.~Costaridou, ``Assessing heterogeneity of lesion enhancement kinetics in
  dynamic contrast-enhanced {MRI} for breast cancer diagnosis,'' \emph{The
  British Journal of Radiology}, 2014.

\bibitem{o2011dynamic}
J.~O\'Connor, P.~Tofts, K.~Miles, L.~Parkes, G.~Thompson, and A.~Jackson,
  ``Dynamic contrast-enhanced imaging techniques: {CT} and {MRI},'' \emph{The
  British Journal of Radiology}, vol.~84.

\bibitem{saha2017effects}
A.~Saha, X.~Yu, D.~Sahoo, and M.~A. Mazurowski, ``Effects of {MRI} scanner
  parameters on breast cancer radiomics,'' \emph{Expert Systems with
  Applications}, vol.~87, pp. 384--391, 2017.

\bibitem{nyul1999standardizing}
L.~G. Ny{\'u}l, J.~K. Udupa \emph{et~al.}, ``On standardizing the {MR} image
  intensity scale,'' \emph{Image}, vol. 1081, 1999.

\bibitem{collewet2004influence}
G.~Collewet, M.~Strzelecki, and F.~Mariette, ``Influence of {MRI} acquisition
  protocols and image intensity normalization methods on texture
  classification,'' \emph{Magnetic Resonance Imaging}, vol.~22, no.~1, pp.
  81--91, 2004.

\bibitem{studholme2004accurate}
C.~Studholme, V.~Cardenas, E.~Song, F.~Ezekiel, A.~Maudsley, and M.~Weiner,
  ``Accurate template-based correction of brain {MRI} intensity distortion with
  application to dementia and aging,'' \emph{IEEE Transactions on Medical
  Imaging}, vol.~23, no.~1, pp. 99--110, 2004.

\bibitem{vovk2007review}
``A review of methods for correction of intensity inhomogeneity in {MRI},
  author={Vovk, Uro and Pernus, Franjo and Likar, Botjan}, journal={IEEE
  Transactions on Medical Imaging}, volume={26}, number={3}, pages={405--421},
  year={2007}, publisher={IEEE}.''

\bibitem{weisenfeld2004normalization}
N.~L. Weisenfeld and S.~Warfteld, ``Normalization of joint image-intensity
  statistics in {MRI} using the {K}ullback-{L}eibler divergence,'' in
  \emph{Biomedical Imaging: Nano to Macro, 2004. IEEE International Symposium
  on}.\hskip 1em plus 0.5em minus 0.4em\relax IEEE, 2004, pp. 101--104.

\bibitem{zhang2017detecting}
J.~Zhang, M.~Liu, and D.~Shen, ``Detecting anatomical landmarks from limited
  medical imaging data using two-stage task-oriented deep neural networks,''
  \emph{IEEE Transactions on Image Processing}, vol.~26, no.~10, pp.
  4753--4764, 2017.

\bibitem{cao2018deformable}
X.~Cao, J.~Yang, J.~Zhang, Q.~Wang, P.-T. Yap, and D.~Shen, ``Deformable image
  registration using cue-aware deep regression network,'' \emph{IEEE
  Transactions on Biomedical Engineering}, 2018.

\bibitem{zhang2017joint}
J.~Zhang, M.~Liu, L.~Wang, S.~Chen, P.~Yuan, J.~Li, S.~G.-F. Shen, Z.~Tang,
  K.-C. Chen, J.~J. Xia, and D.~Shen, ``Joint craniomaxillofacial bone
  segmentation and landmark digitization by context-guided fully convolutional
  networks,'' in \emph{International Conference on Medical Image Computing and
  Computer-Assisted Intervention}.\hskip 1em plus 0.5em minus 0.4em\relax
  Springer, 2017, pp. 720--728.

\bibitem{pereira2016brain}
S.~Pereira, A.~Pinto, V.~Alves, and C.~A. Silva, ``Brain tumor segmentation
  using convolutional neural networks in {MR} images,'' \emph{IEEE Transactions
  on Medical Imaging}, vol.~35, no.~5, pp. 1240--1251, 2016.

\bibitem{lian2018multi}
C.~Lian, J.~Zhang, M.~Liu, X.~Zong, S.-C. Hung, W.~Lin, and D.~Shen,
  ``Multi-channel multi-scale fully convolutional network for 3{D} perivascular
  spaces segmentation in 7{T} {MR} images,'' \emph{Medical Image Analysis},
  vol.~46, pp. 106--117, 2018.

\bibitem{liu2018landmark}
M.~Liu, J.~Zhang, E.~Adeli, and D.~Shen, ``Landmark-based deep multi-instance
  learning for brain disease diagnosis,'' \emph{Medical Image Analysis},
  vol.~43, pp. 157--168, 2018.

\bibitem{xu2017gland}
Y.~Xu, Y.~Li, Y.~Wang, M.~Liu, Y.~Fan, M.~Lai, I.~Eric, and C.~Chang, ``Gland
  instance segmentation using deep multichannel neural networks,'' \emph{IEEE
  Transactions on Biomedical Engineering}, vol.~64, no.~12, pp. 2901--2912,
  2017.

\bibitem{bi2017dermoscopic}
L.~Bi, J.~Kim, E.~Ahn, A.~Kumar, M.~Fulham, and D.~Feng, ``Dermoscopic image
  segmentation via multistage fully convolutional networks,'' \emph{IEEE
  Transactions on Biomedical Engineering}, vol.~64, no.~9, pp. 2065--2074,
  2017.

\bibitem{yap2017automated}
M.~H. Yap, G.~Pons, J.~Mart{\'\i}, S.~Ganau, M.~Sent{\'\i}s, R.~Zwiggelaar,
  A.~K. Davison, and R.~Mart{\'\i}, ``Automated breast ultrasound lesions
  detection using convolutional neural networks,'' \emph{IEEE Journal of
  Biomedical and Health Informatics}, 2017.

\bibitem{ronneberger2015u}
O.~Ronneberger, P.~Fischer, and T.~Brox, ``U-net: {C}onvolutional networks for
  biomedical image segmentation,'' in \emph{International Conference on Medical
  Image Computing and Computer-Assisted Intervention}.\hskip 1em plus 0.5em
  minus 0.4em\relax Springer, 2015, pp. 234--241.

\bibitem{grimm2015computational}
L.~J. Grimm, J.~Zhang, and M.~A. Mazurowski, ``Computational approach to
  radiogenomics of breast cancer: {L}uminal {A} and luminal {B} molecular
  subtypes are associated with imaging features on routine breast {MRI}
  extracted using computer vision algorithms,'' \emph{Journal of Magnetic
  Resonance Imaging}, vol.~42, no.~4, pp. 902--907, 2015.

\bibitem{arasu2011can}
V.~A. Arasu, R.~C. Chen, D.~N. Newitt, C.~B. Chang, H.~Tso, N.~M. Hylton, and
  B.~N. Joe, ``Can signal enhancement ratio ({SER}) reduce the number of
  recommended biopsies without affecting cancer yield in occult {MRI}-detected
  lesions?'' \emph{Academic Radiology}, vol.~18, no.~6, pp. 716--721, 2011.

\bibitem{wang2015identifying}
J.~Wang, F.~Kato, N.~Oyama-Manabe, R.~Li, Y.~Cui, K.~K. Tha, H.~Yamashita,
  K.~Kudo, and H.~Shirato, ``Identifying triple-negative breast cancer using
  background parenchymal enhancement heterogeneity on dynamic contrast-enhanced
  {MRI}: {A} pilot radiomics study,'' \emph{PLoS One}, vol.~10, no.~11, p.
  e0143308, 2015.

\bibitem{wan2016radio}
T.~Wan, B.~N. Bloch, D.~Plecha, C.~L. Thompson, H.~Gilmore, C.~Jaffe,
  L.~Harris, and A.~Madabhushi, ``A radio-genomics approach for identifying
  high risk estrogen receptor-positive breast cancers on {DCE-MRI}:
  {P}reliminary results in predicting {O}ncotype{DX} risk scores,''
  \emph{Scientific Reports}, vol.~6, p. 21394, 2016.

\bibitem{haralick1973textural}
R.~M. Haralick, K.~Shanmugam \emph{et~al.}, ``Textural features for image
  classification,'' \emph{IEEE Transactions on Systems, Man, and Cybernetics},
  no.~6, pp. 610--621, 1973.

\bibitem{zhang2016detecting}
J.~Zhang, Y.~Gao, Y.~Gao, B.~C. Munsell, and D.~Shen, ``Detecting anatomical
  landmarks for fast {A}lzheimer's disease diagnosis,'' \emph{IEEE Transactions
  on Medical Imaging}, vol.~35, no.~12, pp. 2524--2533, 2016.

\end{thebibliography}
\end{document}